\documentclass[runningheads]{llncs}
\usepackage{graphicx}
\usepackage{amsmath}
\usepackage{amssymb}
\usepackage{array}
\usepackage{algorithm}  
\usepackage{algpseudocode}  
\usepackage{subfigure}
\usepackage{color}
\usepackage[misc]{ifsym}

\begin{document}
\title{Robust Ensembling Network for Unsupervised Domain Adaptation}
%
%\titlerunning{Abbreviated paper title}
% If the paper title is too long for the running head, you can set
% an abbreviated paper title here
%
\title{Robust Ensembling Network for Unsupervised Domain Adaptation}

\author{Han Sun\textsuperscript{1(\Letter)}\and
	Lei Lin\textsuperscript{1} \and Ningzhong Liu\textsuperscript{1} \and
	Huiyu Zhou\textsuperscript{2}}

\authorrunning{H. Sun et al.}

\institute{\textsuperscript{1} Nanjing University of Aeronautics and Astronautics, Jiangsu Nanjing, China\\
	\textsuperscript{2} School of Informatics, University of Leicester, Leicester LE1 7RH, U.K\\
	\email{sunhan@nuaa.edu.cn}	
}
\maketitle              % typeset the header of the contribution
\begin{abstract}
Recently, in order to address the unsupervised domain adaptation (UDA) problem, extensive studies have been proposed to achieve transferrable models. Among them, the most prevalent method is adversarial domain adaptation, which can shorten the distance between the source domain and the target domain. Although adversarial learning is very effective, it still leads to the instability of the network and the drawbacks of confusing category information. In this paper, we propose a Robust Ensembling Network (REN) for UDA, which applies a robust time ensembling teacher network to learn global information for domain transfer. Specifically, REN mainly includes a teacher network and a student network, which performs standard domain adaptation training and updates weights of the teacher network. In addition, we also propose a dual-network conditional adversarial loss to improve the ability of the discriminator. Finally, for the purpose of improving the basic ability of the student network, we utilize the consistency constraint to balance the error between the student network and the teacher network. Extensive experimental results on several UDA datasets have demonstrated the effectiveness of our model by comparing with other state-of-the-art UDA algorithms.

\end{abstract}
\section{Introduction}
In recent years, deep neural networks have played a particularly critical role in the face of many computer vision tasks, such as image classification \cite{a1}, object detection \cite{a2}, semantic segmentation \cite{a3} and so on. However, training a perfect neural network demands a large amount of data and corresponding data labeling, which is very time-consuming and expensive. When faceing a new task or a new dataset, the previously trained model may exhibit poor performance due to the domain shift. We hope to use available network and data to complete the target task through knowledge transfer, where domain adaptation methods are needed. The problem that domain adaptation figures out is how to adapt the model trained in the source domain with rich labels to the target domain with sparse labels, and minimize the negative transfer. Besides, UDA means that we only have annotations on the source domain data, without the target domain.
\begin{figure}[t]
	\centering
	\includegraphics[width=12cm]{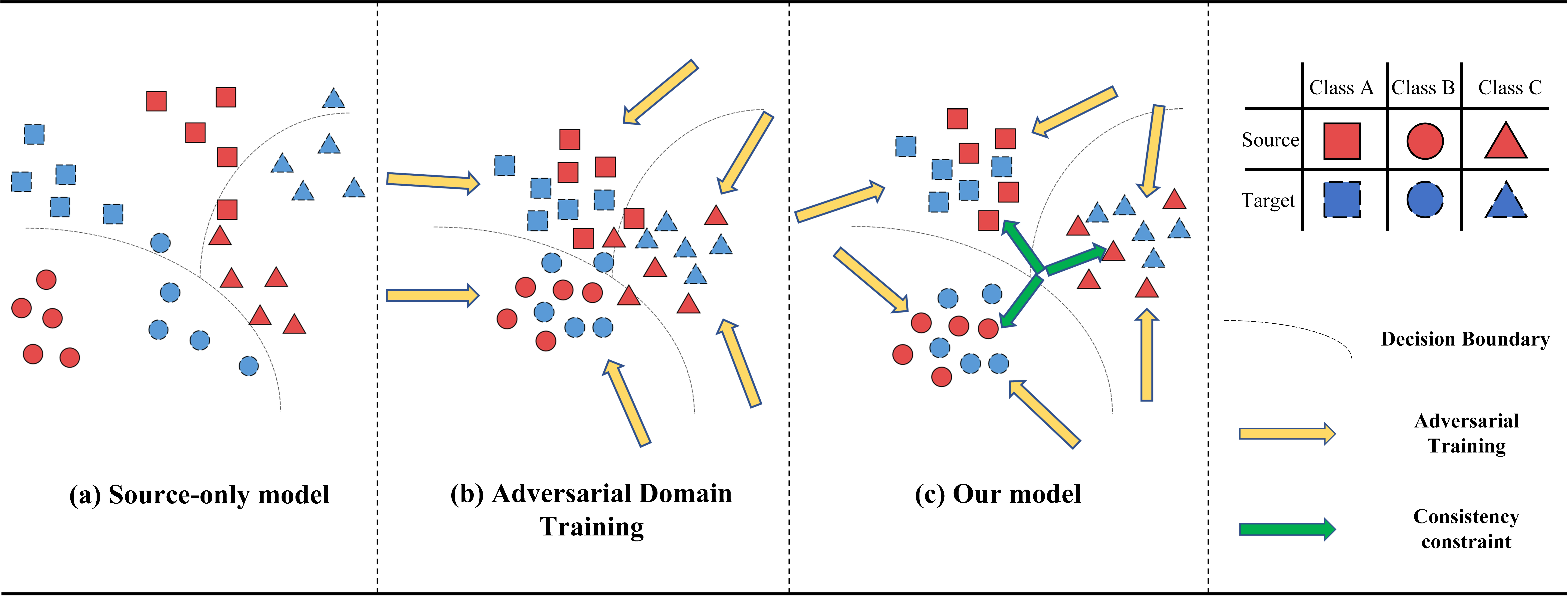}
	\caption{Comparison between previous adversarial domain training method and ours. \textbf{(a):} Before domain adaptation, the data distribution of the source domain and the target domain are quite different. \textbf{(b):} The adversarial domain training method aligns the source domain and the target domain by a domain discriminator, and make them as close as possible to each other, which causes the confusion of the category information. \textbf{(c):} Our model uses the time ensembling algorithm to obtain a more stable network with the consistency constraint, which effectively avoids the distribution of data samples near the category decision boundary.}
	\label{fig1}
\end{figure}

Due to the rapid development of deep learning, many approaches of deep domain adaptation have sprung up \cite{a7}\cite{a6}\cite{TCA}. Among them, many methods try to map the source domain and target domain features into a high-dimensional space, and then perform feature alignment in this space for the reduction of domain shift. The feature alignment method generally uses Maximum Mean Discrepancy (MMD) \cite{MMD} or its improved versions. Subsequently, as a result of the bloom of Generative Adversarial Networks (GAN) \cite{GAN}, the concept of adversarial learning has also been widely employed to domain adaptation. 
%The essential method of adversarial domain adaptation is to train a domain discriminator, the function of which is to differentiate whether the data sample comes from the source domain or the target domain. 
%During overall training the feature extractor should minimize the classification loss of the source domain, and it also maximizes the domain confusion loss of the discriminator. 
In this way, the domain bias between the source domain and the target domain can be effectively reduced by adversarial learning.

The previous UDA methods such as feature alignment, adversarial learning, or clustering methods to generate pseudo-labels for the target domain have been relatively mature, whereas most methods rely on the features or the predictions obtained during the network training process with loss constraint. However, it is assumed that the network with the poor generalization ability itself is unstable during the training process due to the insufficient data, which can cause greater errors in the distribution of the source and target domain features extracted by the network, and may eventually result in disappointing performance and poor robustness of the network. Therefore, it is particularly significant to train a more stable network. In addition, the main popular methods often use adversarial learning. Although adversarial learning can effectively shorten the distance between the source domain and the target domain, it can also confuse the category information between the domains. As shown in Fig. \ref{fig1}, the source domain and target domain samples get closer to each other in the process of adversarial learning, but they also become closer to the decision line, resulting in inaccurate classification. Consequently, we propose a more robust network, and  decrease misclassification through consistency constraint.

In this paper, we explore a robust ensembling network for UDA, which captures more information-rich global features through a more stable model to achieve domain transfer. Specifically, a basic student network is applied for regular domain adaptation training, and then another ensembling teacher network is applied. The weight of the teacher network is the time series ensembling of the basic student network weights, so that the teacher network not only becames more stable, but also has more global information. The ensembling teacher network is adopted to reversely guide the basic student network to enhance the accuracy of its intra-domain classification. Besides, the instance feature is combined with the prediction of the ensembling teacher network and student network as a new condition for the domain discriminator, thereby adversarially decreasing the difference between domains.
The main contribution points of this article are:
\begin{itemize}
	\item [$\bullet$]This paper proposes a robust ensembling network for UDA, which can reduce the prediction error caused by network fluctuations during the training process. The features and predictions obtained by network during training will be more stable with more global information, and more conducive to domain transfer.
\end{itemize}
\begin{itemize}
	\item [$\bullet$]We employ the predictions of the ensembling teacher network to reversely constrain the basic student network to raise the stability of the basic student network. Besides, we use the new predictions to constitute dual-network conditional adversarial loss and effectively alleviate the phenomenon of negative transfer.
\end{itemize}
\begin{itemize}
	\item [$\bullet$]The proposed network is better than its baseline CDAN\cite{CDAN}, and it presents a competitive result on the various UDA datasets.
\end{itemize}
\section{Related Work}
UDA has been widely studied in computer vision mainly for classification and detection tasks. In the era of deep neural networks, the main idea of domain adaptation is to learn domain invariant features between the source and target domain. Among them, several methods exploit MMD \cite{MMD} and its kernel variants to minimize the difference in feature distribution. 
%The MMD is firstly applied by \cite{TCA} to learn a transferable component in a Reproducing Kernel Hilbert Space. 
With the rise of neural networks, \cite{ghifary2014domain} attempts to introduce MMD as a regularization method to minimize the distribution mismatch between the source and target domain in the latent space. In addition to considering the adaptive algorithm of multi-feature representation, \cite{zhu2019multi} also  provides an improved conditional maximum mean error. 

Recently, adversarial learning-based methods exert a tremendous fascination to bridge the gap between the source domain and the target domain.  GAN \cite{GAN} is motivated by the idea of two-play game in game theory. Adversarial training is the process in which the generator and the discriminator compete against each other. 
%However, in the setting of adversarial domain adaptation, due to the existence of the source domain and the target domain, the two naturally form a process of confrontation, which saves the part of the generator. 
Adversarial learning is firstly applied to domain adaptation in \cite{DANN}. Its core idea is to adopt the discriminator to learn domain invariant features. A more general framework is proposed by \cite{ADDA} for adversarial domain adaptation. The author of \cite{CDAN} is motivated by \cite{CGAN} and proposes to align category labels by using the joint distribution of features and predictions. 
%It is believed by \cite{wang2019transferable} that in the process of adversarial learning, not all images are suitable for transfer. The attention mechanism is used to filter transferable samples. 
Inspired by \cite{arjovsky2017wasserstein},  \cite{shen2018wasserstein} applies the Wasserstein GAN measurement to domain adaptation, and proposes Wasserstein distance guided representation learning.

Semi-supervised learning uses both labeled data and unlabeled data during training. The domain adaptation problem is similar to semi-supervised learning in the strict sense, but the source domain and target domain have domain shift due to various image capture devices, environmental changes, and different styles. 
%Semi-supervised learning methods are mainly divided into the use of consistent regularization methods and pseudo-label \cite{lee2013pseudo} methods to lessen the phenomenon of model overfitting. 
Initially, the application of a time series ensembling is proposed by \cite{laine2016temporal}, which adopts the average of the current model prediction results and the historical prediction results to calculate the mean square error. Different from historically weighted sum of model predictions in \cite{laine2016temporal} , \cite{tarvainen2017mean} uses weighted exponential moving average (EMA) on the weight of the student model. Temporal Ensembling \cite{laine2016temporal} is applied by \cite{french2017self} to the domain adaptation problem, and data augmentation is implemented to increase the generalization ability of the model. The prediction results of time ensembling is utilized as pseudo-labels to cluster and align the feature spaces of the source domain and the target domain in \cite{CAT}.  

With the current adversarial domain adaptation methods, the network tends to overlap the source domain and target domain's category distributions during adversarial training, resulting in poor classification results. However, the consistency constraint of semi-supervised learning can just constrain sample distribution and decrease classification errors. In addition, most of the current semi-supervised learning methods in domain adaptation exploit pseudo-label methods, but the wrong pseudo-label may mislead the training of the network. Moreover, most of the previous methods only adopt prediction ensembling methods, and there are few researches on the model ensembling of \cite{tarvainen2017mean}.  Therefore, based on CDAN \cite{CDAN}, we propose an UDA method of robust ensembling model.

\section{Methodology}
This part we mainly provide the specific steps of the method proposed in this article. First we introduce the overall structure of the method in this article; then we dive deep into the loss function of each part; finally we examines the total loss function of the network.
\begin{figure*}[ht]
	\centering
	\includegraphics[width=12cm]{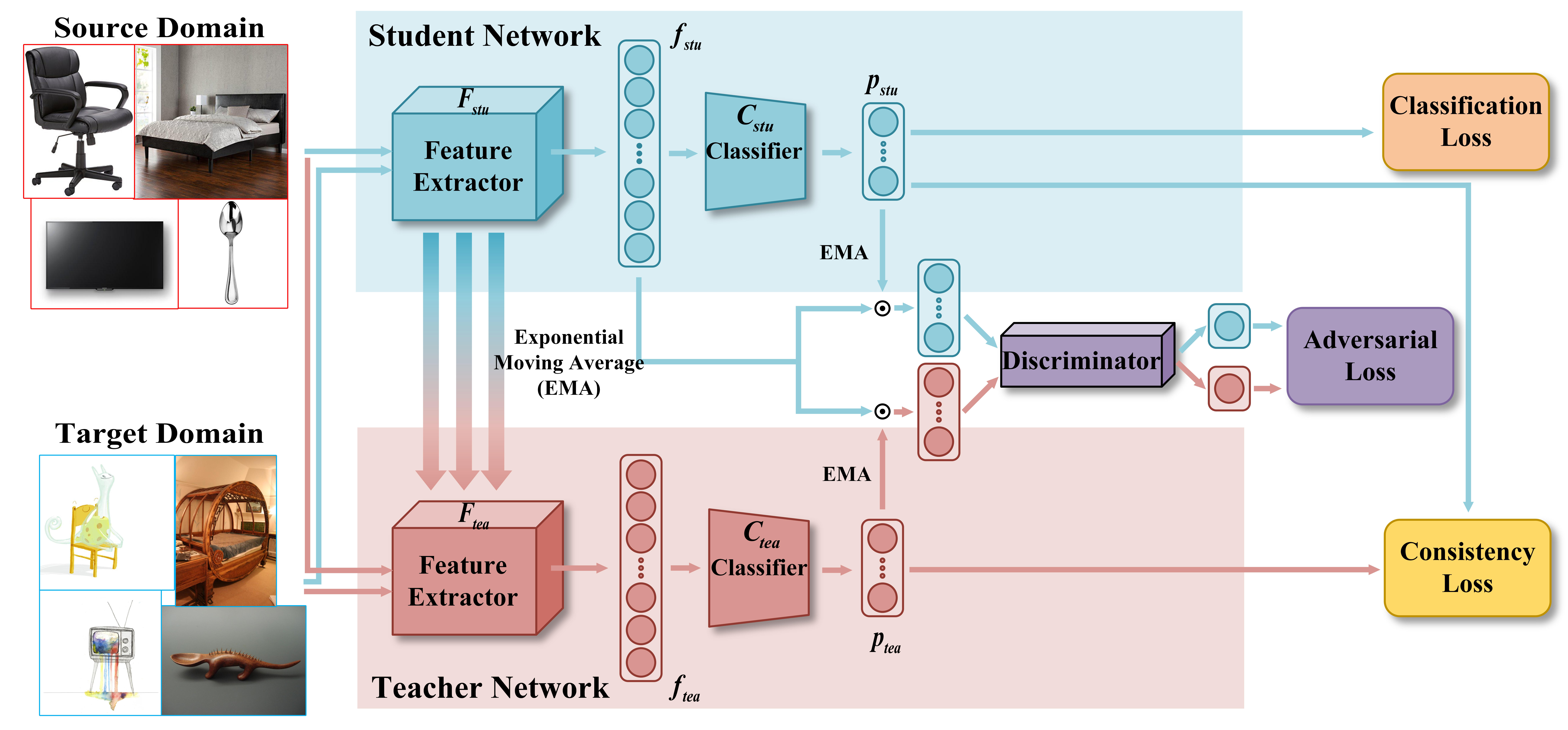}
	\caption{Overview of the proposed REN model. It principally comprises a student network, a teacher network and a domain discriminator. Both the student network and the teacher network consist of a feature extractor and a classifier. The source and target domain samples are delivered to the student network at the same time to extract features, and the weight of the teacher network is the ensembling of the student network's weights in time series. Then, the predictions of the student and the teacher network are ensembled and multiplied with the features of the student network to form a dual-network conditional adversarial loss. Finally, the classification results of the two models are considered together for consistency loss.}
	\label{fig2}
\end{figure*}

\subsection{Overview}
In the UDA problem, the labeled source domain is denoted as $\mathbf{D}_{s}=\left\{\left(\mathbf{x}_{i}^{s}, y_{i}^{s}\right)\right\}_{i=0}^{N_{s}}$, where $\mathbf{N_{s}}$ represents the number of samples with labels in the source domain, and the unlabeled target domain is denoted as $\mathbf{D}_{t}=\left\{\left(\mathbf{x}_{j}^{t}\right)\right\}_{j=0}^{N_{t}}$, where $\mathbf{N_{t}}$ represents the number of samples without labels in the target domain  \cite{Survey}. The source domain and target domain samples conform to the joint distribution $P_{s}\left(X^{s}, Y^{s}\right)$ and $Q_{t}\left(X^{t}, Y^{t}\right)$ respectively, and $P\ne Q$. Our goal is to train a deep neural network $\mathcal{F}:X^t\rightarrow Y^t$ using source domain data with labels and target domain data, which can accurately predict target domain samples while minimizing domain shift.

As shown in Fig. \ref{fig2}, our model mainly includes two networks, a student network and a teacher network, and a domain discriminator $D$. Each network has a feature extractor $F$ and a classifier $C$. Given a picture $x$, the corresponding feature vector $f=F\left(x\right)\in\mathbb{R}^d$ is obtained through the feature extractor $F$, where $d$ represents the feature dimension, and the corresponding prediction result $p=C\left(f\right)\in\mathbb{R}^c$ is obtained through the classifier $C$, where $c$ represents the total number of classes \cite{DANN}. According to \cite{DANN}, the adversarial domain adaptation method can be expressed as optimizing the following minimum and maximum problem:

\begin{equation}
	\min _{F, C} \max _{D} \mathcal{L}_{c}(F, C)-\lambda_{d} \mathcal{L}_{d}(F, D),
\end{equation}
\begin{equation}
	\mathcal{L}_{c}(F, C)=-\mathbb{E}_{\left(x_{s}, y_{s}\right) \sim\left(X_{s}, Y_{s}\right)} \sum_{n=1}^{N_{s}}\left[\mathbf{1}_{\left[n=y_{s}\right]} \log \left[C\left(F\left(x_{s}\right)\right)\right]\right],
\end{equation}
\begin{equation}
	\mathcal{L}_{d}(F, D)=-\mathbb{E}_{x_{s} \sim P_{s}} \log \left[D\left(f_{s}\right)\right]-\mathbb{E}_{x_{t} \sim Q_{t}} \log \left[1-D\left(f_{t}\right)\right]\odot,
\end{equation}
Among them, $\mathcal{L}_c\left(F,C\right)$ is the standard supervised classification task, which uses only the cross-entropy loss of labeled source domain data. The domain discriminator $D$ is a two-classifier $D:\mathbb{R}^d\rightarrow[0,1]$, which predicts whether the data comes from the source domain or the target domain.

\subsection{Robust Ensembling Network}

The previous domain adaptation adversarial methods are limited to some extent due to the problem that the model is unrobust caused by adversarial learning. Although adversarial learning effectively narrows the distribution of the source and target domains, it also confuses category information. The semi-supervised learning method is mainly dedicated to finding the optimal classification decision line. 
%The key of these methods is to utilize unlabeled data to strengthen the training model, that is, if different perturbations are implemented during model training, the prediction results should not differ significantly, and under the clustering assumption, data points with different labels should be separated from each other in the low-density area. 
The key of these methods is to enhance the training model with unlabeled data and cluster data points of different labels with perturbations.
%Temporal Ensembling \cite{laine2016temporal} is improved with a time series combination model, using the average of the current model prediction results and the historical prediction results to calculate the mean square error. 
%which can effectively retain historical information, eliminate disturbances and stabilize the current value. 
%Subsequently, Mean Teacher \cite{tarvainen2017mean} thinks it's better to integrate the entire network with the time series, which would alleviate the space overhead and avoid the lag of the Temporal Ensembling \cite{laine2016temporal} model.

We believe that semi-supervised learning and domain adaptation have something in common. 
%The source domain data is labeled data in semi-supervised learning, and the target domain is similar to unlabeled data. 
Both semi-supervised learning and UDA are limited to the fact that part of the data is not labeled, which makes their solutions intersect.
The only difference is that there is a distribution difference between the source domain and the target domain in UDA, so semi-supervised learning methods can be applied to enhance the robustness of the model and raise model prediction accuracy. The main method used in this paper is the mean teacher method, which contains a student network and a teacher network. The student network is trained normally, and the weight update of the teacher network is integrated by the time of the weight of the student network, and the update method is exponential moving average (EMA) :
\begin{equation}
\theta_n^\prime={\alpha_\theta\theta}_{n-1}^\prime+\left(1-\alpha_\theta\right)\theta_n,
\end{equation}
Among them, $\theta_n$ represents the weight of the student model during the $n$-th training, $\theta_n^\prime$ represents the weight of the teacher model during the $n$-th training, and $\alpha_\theta$ is the smoothing coefficient hyperparameter of the network parameter.

\subsection{Dual-Network Conditional Adversarial Learning}
The previous adversarial learning is generally limited to the labels without the target domain, so CDAN \cite{CDAN} proposes to jointly consider the feature representation and the prediction result of the classifier in the discriminator part. CDAN believes that the outer product of the feature representation and the prediction result can affect the feature representation, so a discriminator shared by the source domain and the target domain is used to align this conditional feature representation. The general conditional adversarial loss are as follows:
\begin{equation}
	\begin{aligned}
		\mathcal{L}_d^{con}\left(F,D\right)=-\mathbb{E}_{x_s~P_s}\log{\left[D\left(f_s,p_s\right)\right]}-\mathbb{E}_{x_t~Q_t}\log{\left[1-D\left(f_t,p_t\right)\right]}
	\end{aligned}.
\end{equation}

This conditional adversarial loss generally utilizes the feature $f_s$ and prediction $p_s$ of the source domain, and also exploits the feature $f_t$ and prediction $p_t$ of the target domain. However, the features and predictions that this adversarial loss mainly relies on are not stable, and the two may have numerical deviations during the network training process, which will eventually lead to the poor effect of adversarial learning. The dual network proposed in this article can effectively avoid this problem. The student network performs standard domain adaptation adversarial learning, while the update of the teacher network is provided by the student network. The update method is the former time ensembling, so that the teacher network integrates the student network in time series, and the network structure is more stable and robust. In addition to the student's condition adversarial loss, the prediction of the teacher network is also applied as another condition.

\begin{equation}
	\begin{aligned}
		\mathcal{L}_d^{stu}\left(F_{stu},D\right)=-\mathbb{E}_{x_s~P_s}\log{\left[D\left(f_s^{stu},\widehat{p_s^{stu}}\right)\right]}\\
		-\mathbb{E}_{x_t~Q_t}\log{\left[1-D\left(f_t^{stu},\widehat{p_t^{stu}}\right)\right]},
	\end{aligned}	
\end{equation}
\begin{equation}
	\begin{aligned}
		\mathcal{L}_d^{tea}\left(F_{stu},D\right)=-\mathbb{E}_{x_s~P_s}\log{\left[D\left(f_s^{stu},\widehat{p_s^{tea}}\right)\right]}\\
		-\mathbb{E}_{x_t~Q_t}\log{\left[1-D\left(f_t^{stu},\widehat{p_t^{tea}}\right)\right]},
	\end{aligned}	
\end{equation}
\begin{equation}
	\widehat{p_n}=\left(1-\alpha_p\right)\\
	\widehat{p_{n-1}}+ \alpha_p p_n,
\end{equation}
Among them, $p_n$ represents the prediction of the student or teacher model during the $n$-th training, $\widehat{p_n}$ represents the ensembling prediction during the $n$-th training and $\alpha_p$ is the smoothing coefficient hyperparameter. The prediction of the student network and the teacher network performs an EMA operation to promote the stability of the prediction. In this way, with global student predictions and global teacher predictions, the network can learn more reliable conditional and transferable information in the process of adversarial learning.

\subsection{Consistency Constraint}
The student network provides weights for the teacher network, just as student asks the teacher in the classroom, so the teacher needs to answer the student’s questions to help the student. Therefore, the teacher network needs to assist the student network, and the method of assistance is to adopt consistency constraint. The core idea of consistency constraint is to perturb high-dimensional data so that it tends to be consistent in the feature space. In other words, we hope that in the process of dimensionality reduction, multiple high-dimensional data can be compressed into a low-dimensional point, so that the feature distribution in the feature space is more compact, which is conducive to the model learning more accurate classification decision lines. We mainly utilize the L2 norm between the student model and the teacher model as the consistency loss.
\begin{equation}
	\mathcal{L}_{con}=\left \| C_{stu}\left ( F_{stu}\left ( x \right ) \right )-C_{tea}\left ( F_{tea}\left ( x \right ) \right ) \right \|_2 .
\end{equation}

\subsection{Total Loss Function}
In this part, we introduce the total loss function:
\begin{equation}
	\mathcal{L}_{all}=\mathcal{L}_c+{\lambda_d^{stu}\mathcal{L}}_d^{stu}+{\lambda_d^{tea}\mathcal{L}}_d^{tea}+\gamma\mathcal{L}_{con},
\end{equation}
Therefore, the final total loss function is as described above, and it mainly includes four parts: the supervised classification loss $\mathcal{L}_c$ of source domain, a student network adversarial loss $\mathcal{L}_d^{stu}$, a teacher network adversarial loss $\mathcal{L}_d^{tea}$ and finally the consistency loss $\mathcal{L}_{con}$ between the student network and the teacher network. Among them, $\lambda_d^{stu}$ and $\lambda_d^{stu}$ are the hyperparameters of student adversarial loss and teacher adversarial loss, and their role is to control the importance of the two in adversarial training. $\gamma$ is the relative weight that controls the consistency constraint.  %The training process is shown in Algorithm 1.
%  \begin{algorithm}[htb]  
%	\caption{ The optimization strategy of REN.}  
%	\label{alg:Framwork}  
%	\begin{algorithmic}[1]  
%		\Require  
%		Source data as $\mathbf{D}_s=\left\{\left(\mathbf{x}_i^s,y_i^s\right)\right\}_{i=0}^{N_s}$, target data as $\mathbf{D}_t=\left\{\left(\mathbf{x}_j^t\right)\right\}_{j=0}^{N_t}$; The batch-size $n$, and $T$ is training iterations; The student network parameter $\theta_{stu}$ and the teacher $\theta_{tea}$; The student feature extractor $F_{stu}$, the student classifier $C_{stu}$, the teacher feature extractor $F_{tea}$, the teacher classifier $C_{tea}$ and the discriminator $D$;
%		\State  Initialize the student network $\theta_{stu}$ and the teacher network $\theta_{tea}$.
%		\For{$t=1 \to T$} 
%			\State Randomly sample mini-batch of $n$ source samples and $n$ target samples;  
%			\State Obtain features $f_{stu}$ and $f_{tea}$ through $F_{stu}$ and $F_{tea}$, then obtain predictions $p_{stu}$ and $p_{tea}$ through $C_{stu}$ and $C_{tea}$;    
%			\State Update $\widehat{p_{stu}}$ and $\widehat{p_{tea}}$ by Eq.(); 
%			\State Train the discriminator $D$ with $f_{stu}$, $\widehat{p_{stu}}$ and $\widehat{p_{tea}}$ by Eq.(); 
%			\State Train the classifiers $C_{stu}$ with $p_{stu}^s$ and $y^s$ by Eq.() , then train the classifiers $C_{stu}$ and $C_{tea}$ with $p_{stu}$ and $p_{tea}$ by Eq.();  
%			\State Update $\theta_{stu}$ by SGD and $\theta_{tea}$ by Eq.()
%		\EndFor  
%	\end{algorithmic}  
%\end{algorithm}

\section{Experiment}

%In order to prove the effectiveness of the method proposed in this paper, we evaluate our method on the following UDA datasets and compare it with other state-of-the-art methods.
\subsection{Experimental Setting}
\textbf{Datasets.} \textit{Office-31} \cite{office31} contains 31 classes and 4,110 images collected from three differenet domains: Amazon Website (A) with 2817 images, Web Camera (W) with 498 images and Digital SLR Camera (D) with 795 images. By permuting the three domains, we obtain six transfer tasks: A→W, D→W, W→D, A→D, D→A and W→A. 
\textit{ImageCLEF-DA} \cite{JAN} is a dataset created by the ImageCLEF2014 domain adaptation competition. We follow the guidelines of \cite{long2015learning} and select 3 sub-domains of Caltech-256 (C), ImageNet ILSVRC 2012 (I) and Pascal VOC 2012 (P), which have 12 common categories. There are six UDA tasks to be evaluated.
\textit{Office-Home} \cite{office-home} is another more challenging dataset for visual domain adaptation. It mainly includes four dissimilar subdomains, namely Artistic images (Ar), ClipArt (Ca), Product images (Pr) and Real-World images (Re). There are 15500 images in 65 different categories. They are all pictures under office and home settings, which constitute a total of 12 domain adaptation tasks.

%\textbf{VisDA-2017} \cite{visda} is a large computer vision dataset composed of two subdomains: a synthetic dataset, which is mainly composed of 3D model renderings under different angles and different lighting conditions; a real dataset, which is mainly composed of real-world pictures. The dataset is mainly composed of 280k pictures in 12 categories. This scale poses no small challenge to the domain adaptation task. The training images are generated by the same object under different circumstances, and the validation images are collected by MSCOCO \cite{mscoco}.

\textbf{Comparisons.} We compare the REN model with other state-of-the-art methods: (1) ResNet-50 \cite{resnet}. (2) Domain Adversarial Neural Network(DANN) \cite{DANN}. (3) Adversarial Discriminative Domain Adaptation (ADDA) \cite{ADDA}. (4) Deep transfer learning with joint adaptation networks (JAN) \cite{JAN} (5)Conditional Domain Adversarial Network (CDAN) \cite{CDAN}. (6) Cluster Alignment with a Teacher for Unsupervised Domain Adaptation (CAT) \cite{CAT}. (7) Towards Discriminability and Diversity: Batch Nuclear-norm Maximization under Label Insufficient Situations (BNM) \cite{cui2020towards}. 
%To be fair, the results on the Office-31, ImageCLEF-DA, and Office-Home datasets are all derived from these papers if available.

%\begin{table}[t]
%	\centering
%	\caption{Accuracy (\%) on VisDA-2017 for unsupervised domain adaptation(ResNet-50)}
%	\begin{tabular}{ccccccccccccc|c}% 通过添加 | 来表示是否需要绘制竖线
%		\hline  % 在表格最上方绘制横线
%		Method&plane&bcycl&bus
%		&car&horse&knife
%		&mcycl&person&plant
%		&sktbrd&train&truck&Avg\\
%		\hline  %在第一行和第二行之间绘制横线
%		ResNet-50\cite{resnet}&99.9\\
%		DANN\cite{DANN}&99.9\\
%		ADDA\cite{ADDA}&99.9\\
%		CDAN\cite{CDAN}&99.9\\
%		CAT\cite{CAT}&99.9\\
%		ETD\cite{li2020enhanced}&99.9\\
%		BNM\cite{cui2020towards}&99.9\\
%		REN(Ours)&99.9&&&&&&\\
%		\hline % 在表格最下方绘制横线
%	\end{tabular}
%\end{table}

\textbf{Implementation details.} The method proposed in this paper is mainly implemented on the Pytorch framework. For a fair comparison, we apply the same network structure in each experiment. We utilize ResNet50 pre-trained on ImageNet without the final fully connected layer as the feature extractor. We adopt all the labeled data in the source domain and all the unlabeled data in the target domain. We apply the SGD optimizer with a momentum of 0.9, the batch size is 32, and the dimension of the bottleneck layer is set to 256. We adopt the learning rate annealing strategy as \cite{DANN}: the learning rate is adjusted by $\eta_p=\eta_0(1+\alpha p)^{-\beta}$, where $p$ is the training progress changing from 0 to 1, and $\eta_0=0.01$, $\alpha=10$, $\beta=0.75$ are optimized by the importance-weighted cross-validation \cite{sugiyama2007covariate}. In the testing phase, we mainly choose the more stable teacher model for testing.

\subsection{Results}

Table \ref{tab1} shows the UDA results of the six transfer tasks of the Office-31 dataset. We can observe that the performance of the REN method in this paper is much better than all the previous methods on most tasks. It is worth noting that our method REN is not only on simple transfer tasks, such as D→W and W→D, with superior performance, reaching almost 100\% accuracy, but also on tasks that are difficult to transfer due to unbalanced samples, such as D→A and W→A, which have achieved superior results. The main reason for the success of our model is that we have introduced a more robust time ensembling teacher model. The adversarial training of the student model and the teacher model effectively solves the domain offset and enhances the predictive ability of the model.
\begin{table*}[h]
	\centering
	\caption{Accuracy (\%) on Office-31 for UDA (ResNet-50)}
	\setlength{\tabcolsep}{2.5mm}{
		\begin{tabular}{ccccccc|c}% 通过添加 | 来表示是否需要绘制竖线
			\hline  % 在表格最上方绘制横线
			Method&A→W&D→W&W→D&A→D&D→A&W→A&Avg\\
			\hline  %在第一行和第二行之间绘制横线
			ResNet-50\cite{resnet}&68.4&96.7&99.3&68.9&62.5&60.7&76.1\\
			DANN\cite{DANN}&82.0&96.9&99.1&79.7&68.2&67.4&82.2\\
			ADDA\cite{ADDA}&86.2&96.2&98.4&77.8&69.5&68.9&82.9\\
			JAN\cite{JAN}&86.0&96.7&99.7&85.1&69.2&70.7&84.6\\
			CDAN\cite{CDAN}&93.1&98.2&\textbf{100.0}&89.8&70.1&68.0&86.6\\
			CDAN+E\cite{CDAN}&94.1&98.6&\textbf{100.0}&92.9&71.0&69.3&87.7\\
			CAT\cite{CAT}&94.4&98.0&\textbf{100.0}&90.8&72.2&70.2&87.6\\
%			ETD\cite{li2020enhanced}&92.1&\textbf{100.0}&\textbf{100.0}&88.0&71.0&67.8&86.2\\
			BNM\cite{cui2020towards}&92.8&98.8&\textbf{100.0}&92.9&73.5&73.8&88.6\\
			REN(Ours)&\textbf{95.0}&\textbf{99.2}&\textbf{100.0}&\textbf{94.6}&\textbf{74.1}&\textbf{74.8}&\textbf{89.6}\\
			\hline % 在表格最下方绘制横线
			\label{tab1}
	\end{tabular}}
\end{table*}

The results of the six transfer tasks of the ImageCLEF-DA dataset are shown in Table \ref{tab2}. Although the number of images in each subdomain in the ImageCLEF-DA dataset is similar, it is still challenging for the transfer task because of the images from various scenarios. Compared to ResNet-50, which only utilizes source domain samples for fine-tuning, the above-mentioned domain adaptation method achieves significant effect. Compared with other methods, the method in this article has achieved significant improvement. 
The CAT method also adopts the idea of semi-supervised learning, but it only adopts the Pi-model \cite{laine2016temporal} prediction ensembling method, and we utilize model ensembling, which is more effective. This also proves that the student-teacher model of REN can learn more transferable features.
\begin{table*}[ht]
	\centering
	\caption{Accuracy (\%) on ImageCLEF-DA for UDA (ResNet-50)}
	\setlength{\tabcolsep}{3.3mm}{
		\begin{tabular}{ccccccc|c}% 通过添加 | 来表示是否需要绘制竖线
			\hline  % 在表格最上方绘制横线
			Method&I→P&P→I&I→C&C→I&C→P&P→C&Avg\\
			\hline  %在第一行和第二行之间绘制横线
			ResNet-50\cite{resnet}&74.8&83.9&91.5&78.0&65.5&91.2&80.7\\
			DANN\cite{DANN}&75.0&86.0&96.2&87.0&74.3&91.5&85.0\\
			JAN\cite{JAN}&76.8&88.0&94.7&89.5&74.2&91.7&85.8\\
			CDAN\cite{CDAN}&76.7&90.6&97.0&90.5&74.5&93.5&87.1\\
			CDAN+E\cite{CDAN}&77.7&90.7&97.7&91.3&74.2&94.3&87.7\\
			CAT\cite{CAT}&77.2&91.0&95.5&91.3&75.3&93.6&87.3\\
			%ETD\cite{li2020enhanced}&81.0&91.7&97.9&93.3&79.5&95.0&89.7\\
			REN(Ours)&\textbf{79.8}&\textbf{93.3}&\textbf{97.3}&\textbf{91.5}&\textbf{76.8}&\textbf{94.8}&\textbf{88.9}\\
			\hline % 在表格最下方绘制横线
			\label{tab2}
	\end{tabular}}
\end{table*}
\begin{table}[tbp!]
	\centering
	\caption{Accuracy (\%) on Office-Home for UDA (ResNet-50)}
	\resizebox{\textwidth}{!}{
		\begin{tabular}{ccccccccccccc|c}% 通过添加 | 来表示是否需要绘制竖线
%			\hline  % 在表格最上方绘制横线
%			Method&Ar→Cl&Ar→Pr&Ar→Rw
%			&Cl→Ar&Cl→Pr&Cl→Rw
%			&Pr→Ar&Pr→Cl&Pr→Rw
%			&Rw→Ar&Rw→Cl&Rw→Pr&Avg\\
			\hline  %在第一行和第二行之间绘制横线
			&Ar&Ar&Ar
			&Cl&Cl&Cl
			&Pr&Pr&Pr
			&Rw&Rw&Rw&\\

			Method&↓&↓&↓
			&↓&↓&↓
			&↓&↓&↓
			&↓&↓&↓&Avg\\

			&Cl&Pr&Rw
			&Ar&Pr&Rw
			&Ar&Cl&Rw
			&Ar&Cl&Pr&\\
			\hline  %在第一行和第二行之间绘制横线
			ResNet-50\cite{resnet}&34.9&50.0&58.0&37.4&41.9&46.2&38.5&31.2&60.4&53.9&41.2&59.9&46.1\\
			DANN\cite{DANN}&45.6&59.3&70.1&47.0&58.5&60.9&46.1&43.7&68.5&63.2&51.8&76.8&57.6\\
			JAN\cite{JAN}&45.9&61.2&68.9&50.4&59.7&61.0&45.8&43.4&70.3&63.9&52.4&76.8&58.3\\
			CDAN\cite{CDAN}&49.0&69.3&74.5&54.4&66.0&68.4&55.6&48.3&75.9&68.4&55.4&80.5&63.8\\
			CDAN+E\cite{CDAN}&50.7&70.6&76.0&57.6&70.0&70.0&57.4&50.9&77.3&70.9&56.7&81.6&65.8\\
			%			ETD\cite{li2020enhanced}&51.3&71.9&\textbf{85.7}&57.6&69.2&73.7&57.8&51.2&79.3&70.2&57.5&82.1&67.3\\
			BNM\cite{cui2020towards}&52.3&\textbf{73.9}&\textbf{80.0}&\textbf{63.3}&\textbf{72.9}&\textbf{74.9}&\textbf{61.7}&49.5&\textbf{79.7}&70.5&53.6&82.2&67.9\\
			REN(Ours)&\textbf{54.4}&73.6&77.4&61.6&71.1&71.7&61.0&\textbf{52.2}&78.8&\textbf{73.1}&\textbf{59.4}&\textbf{83.5}&\textbf{68.2}\\
			\hline % 在表格最下方绘制横线
			\label{tab3}
	\end{tabular}}
\end{table}

Table \ref{tab3} shows the results of 12 transfer tasks on the Office-Home dataset. Different from the first two datasets, there are more categories in the Office-Home dataset, thus leads to the methods which perform well on the Office-31 dataset may have performance degradation on the Office-Home dataset. Although the method in this paper only has the best effect on 5 transfer tasks, the average accuracy is even better than BNM method. The main reason for this is that the sample size of some categories in Office-Home is extremely unbalanced. For example, the Ruler class in the Art subdomain has only 15 pictures, while the Bottle class has 99 pictures. The main problem that the BNM method solves is this kind of imbalance. Though the method in this paper does not focus on the imbalance of the dataset, due to the stability of the dual network and the ensembling teacher network, the final classification result is not much different from that of BNM. In addition, our method is more effective on difficult tasks such as Ar→Cl, Pr→Cl and Rw→Cl, which are improved by 2.1\%, 2.7\% and 5.8\% respectively than BNM.
The above experiments prove the effectiveness of our method.

%The last experimental result, Table 4, is the dataset VisDA-2017. Although the data set has only one migration task, because of its large scale, there are a total of 180k pictures, and the source domain is composed of 3D renderings, and the target domain is a real picture set. The domain difference between the two is very large, so the VisDA-2017 dataset is extremely challenging.

\subsection{Ablation Study and Visualization}
\begin{table*}[htp!]
	\centering
	\caption{Accuracy (\%) in ablation experiments for REN based on CDAN on Office-31}
	\setlength{\tabcolsep}{1.5mm}{
		\begin{tabular}{ccccccc|c}% 通过添加 | 来表示是否需要绘制竖线
			\hline  % 在表格最上方绘制横线
			Method&A→W&D→W&W→D&A→D&D→A&W→A&Avg\\
			\hline  %在第一行和第二行之间绘制横线
			CDAN&93.1&98.2&\textbf{100.0}&89.8&70.1&68.0&86.6\\
			CDAN+M&93.3&98.9&\textbf{100.0}&92.7&71.5&73.0&88.3\\
			CDAN+M+D&94.8&99.0&\textbf{100.0}&93.7&72.2&74.5&89.0\\
			CDAN+M+D+C(REN)&\textbf{95.0}&\textbf{99.2}&\textbf{100.0}&\textbf{94.6}&\textbf{74.1}&\textbf{74.8}&\textbf{89.6}\\
			\hline % 在表格最下方绘制横线
			\label{tab4}
	\end{tabular}}
\end{table*}

Table \ref{tab4} presents ablation experiments on Office-31 dataset based on CDAN. In this table, we denote performing in mean teacher model ensembling as``M", dual-network conditional adversarial loss as ``D" and consistency constraint as ``C". On Office-31, CDAN+M outperforms CDAN by 1.7\%. In addition, compared with CDAN, CDAN+M+D and CDAN+M+D+C improve its accuracy by 2.4\%, 3\%, indicating the effectiveness of our method. 
\begin{figure}[htp!]
	\centering
	\subfigure[W→A (Office-31)]{
		\centering
		\includegraphics[width=0.47\textwidth]{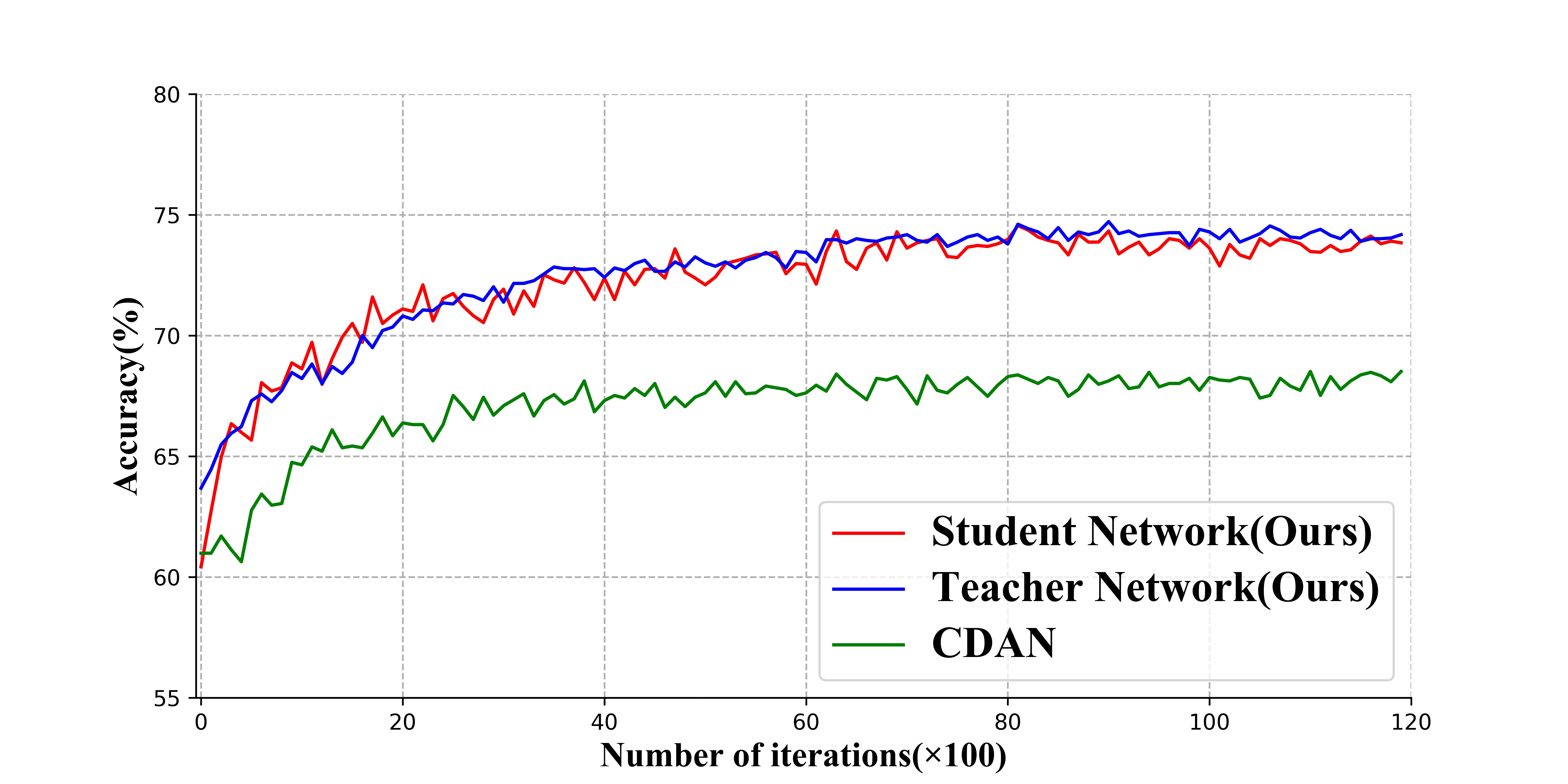}
	}
	\subfigure[P→I (ImageCLEF-DA)]{
		\centering
		\includegraphics[width=0.47\textwidth]{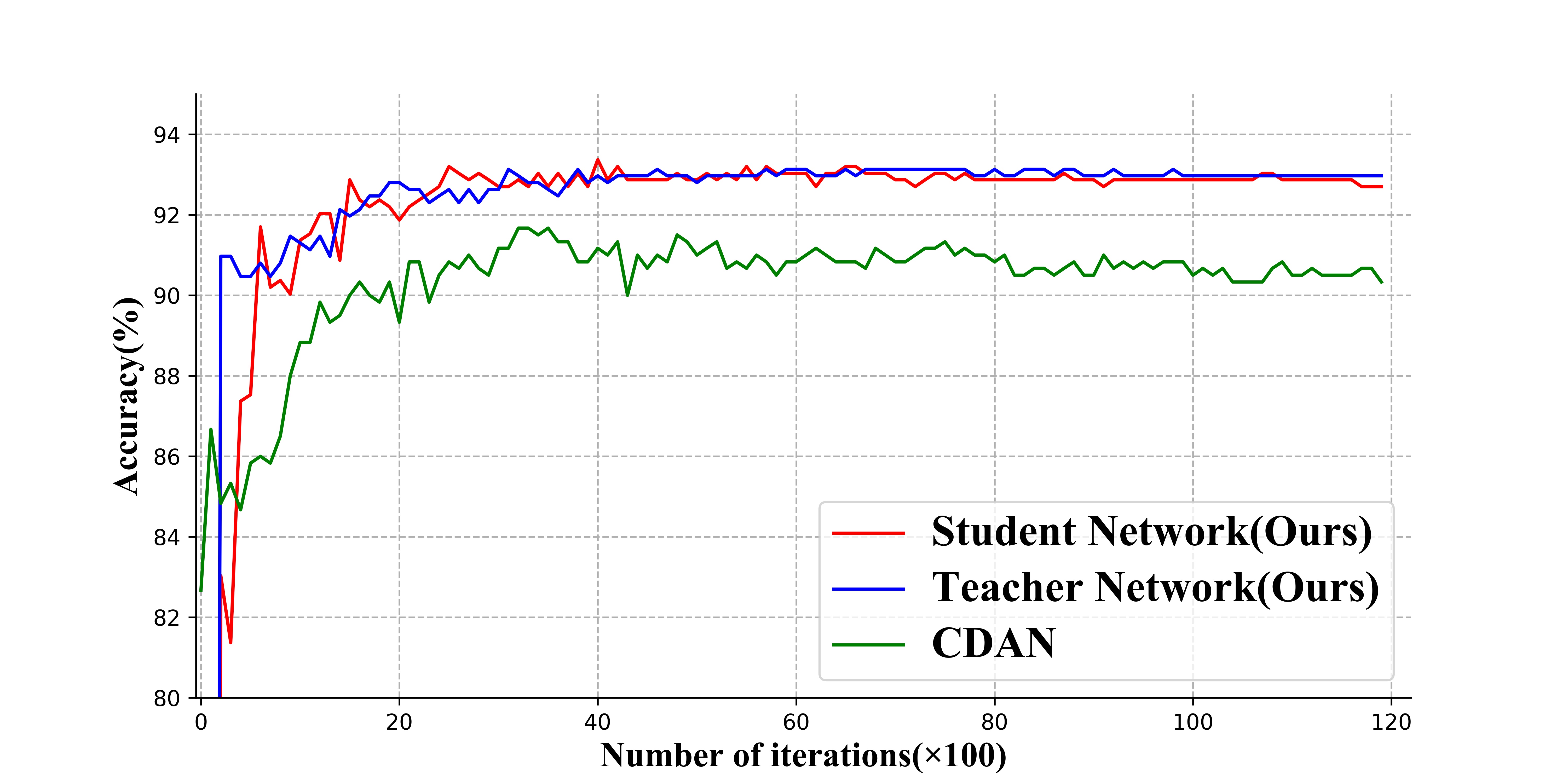}
	}
	\caption{Comparison of stablility of CDAN, student network and teacher network of our model REN. Obviously, the blue line (teacher network) is much smoother and more stable than the red line (student network).}
	\label{fig3}
\end{figure}

In order to demonstrate the stability of the method in the training process, we present the results of the classification accuracy of different training processes in Fig. \ref{fig3}. The graph on the left shows the experimental results of task W→A (Office-31), and the right presents the results of task P→I (ImageCLEF-DA). We can find that the accuracy of both the student network and the teacher network in this article is much higher than the result of CDAN. By observing Fig. \ref{fig3} carefully, we can find that the accuracy curve of the teacher network has less fluctuations than the student network, and the accuracy has also been improved compared with CDAN. In addition, the reason why the accuracy of the student network is also excellent is that the consistency constraint of the teacher network promotes the improvement of the prediction result of the student network. Therefore, according to the curve comparison between CDAN and Teacher Network(Ours), our method has less fluctuation and more stability.

In addition, compared to CDAN, our method is composed of student network and teacher network. It can be seen from Table \ref{tab5} that the parameters of our method is twice that of CDAN during training, and the training time is relatively longer.
\begin{table*}[htp!]
	\centering
	\caption{Comparison of parameters with CDAN in the training phase on Office-31(12k epoches)}
	\setlength{\tabcolsep}{10mm}{
		\begin{tabular}{ccc}% 通过添加 | 来表示是否需要绘制竖线
			\hline  % 在表格最上方绘制横线
			Method&Params&Time\\
			\hline  %在第一行和第二行之间绘制横线
			CDAN&24M&1.3h\\
			REN(Ours)&48M&1.8h\\
			\hline % 在表格最下方绘制横线
			\label{tab5}
	\end{tabular}}
\end{table*}
\subsection{Ablation Study and Visualization}

\begin{figure}[h]
	\centering
	\subfigure[Before training]{
		\centering
		\includegraphics[width=0.3\textwidth]{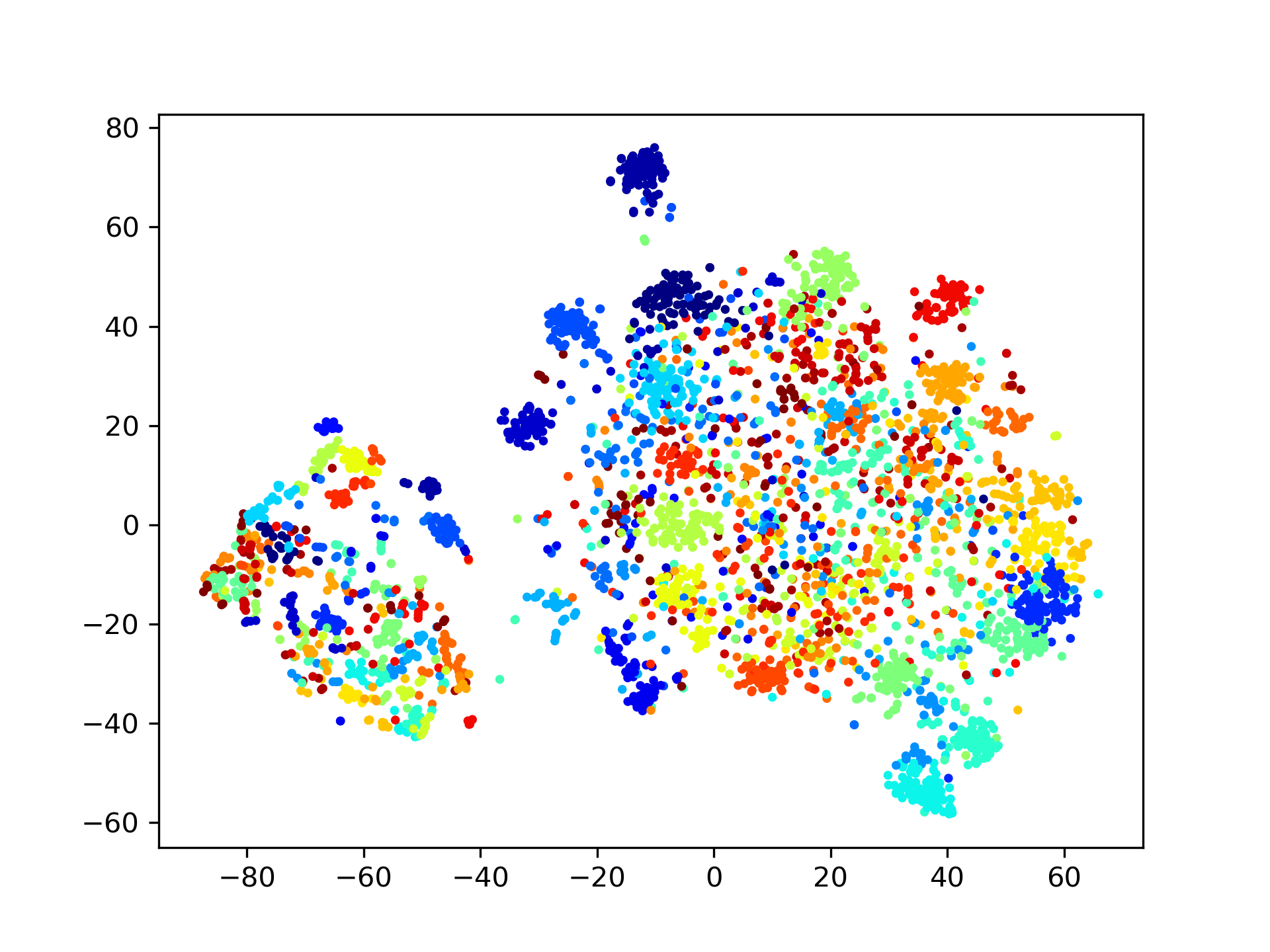}
	}
	%	\subfigure[ResNet-50]{
	%	\centering
	%	\includegraphics[width=0.22\textwidth]{./pic/a2wRC.png}
	%	}
	\subfigure[CDAN]{
		\centering
		\includegraphics[width=0.3\textwidth]{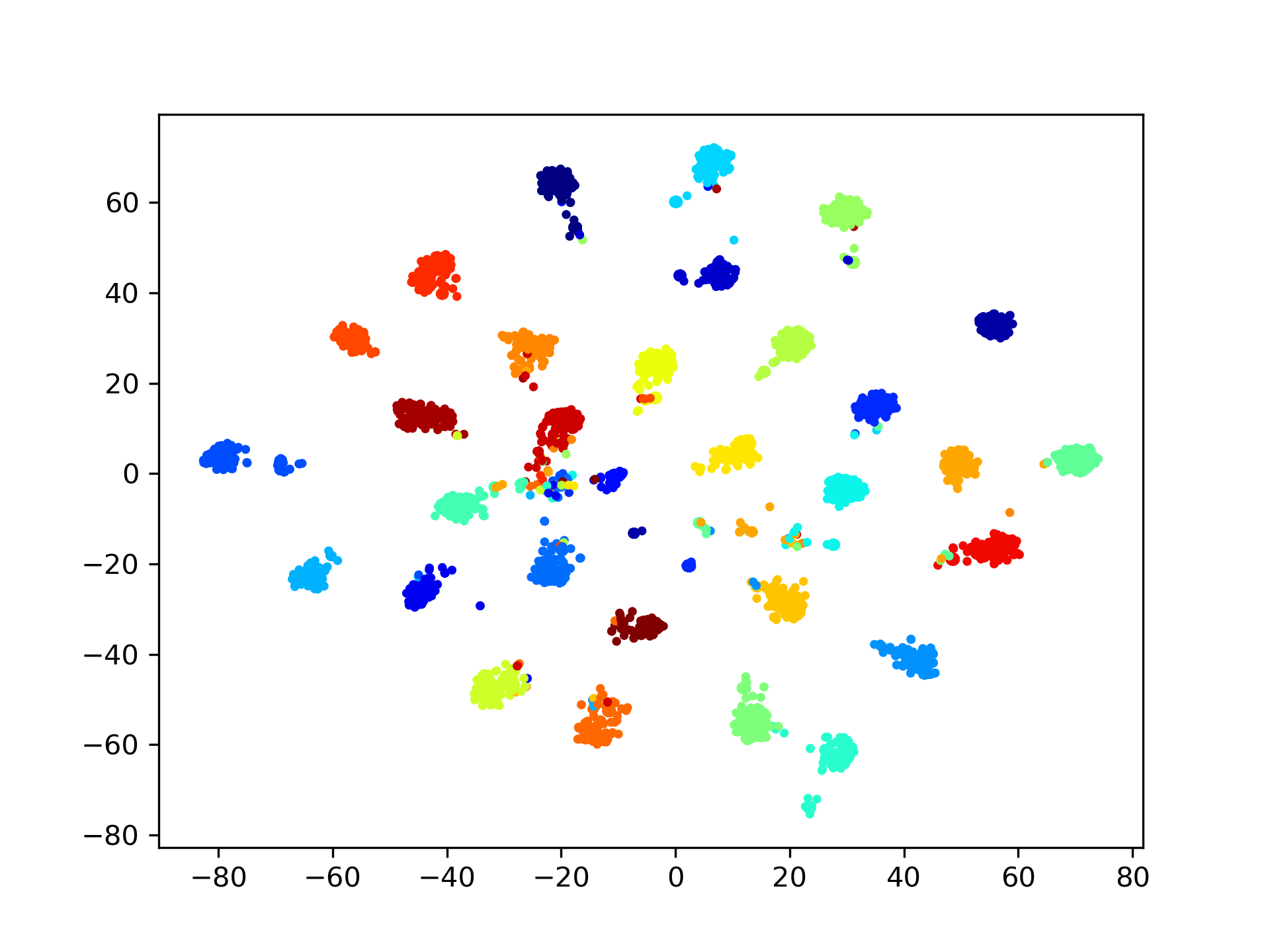}
	}
	\subfigure[REN]{
		\centering
		\includegraphics[width=0.3\textwidth]{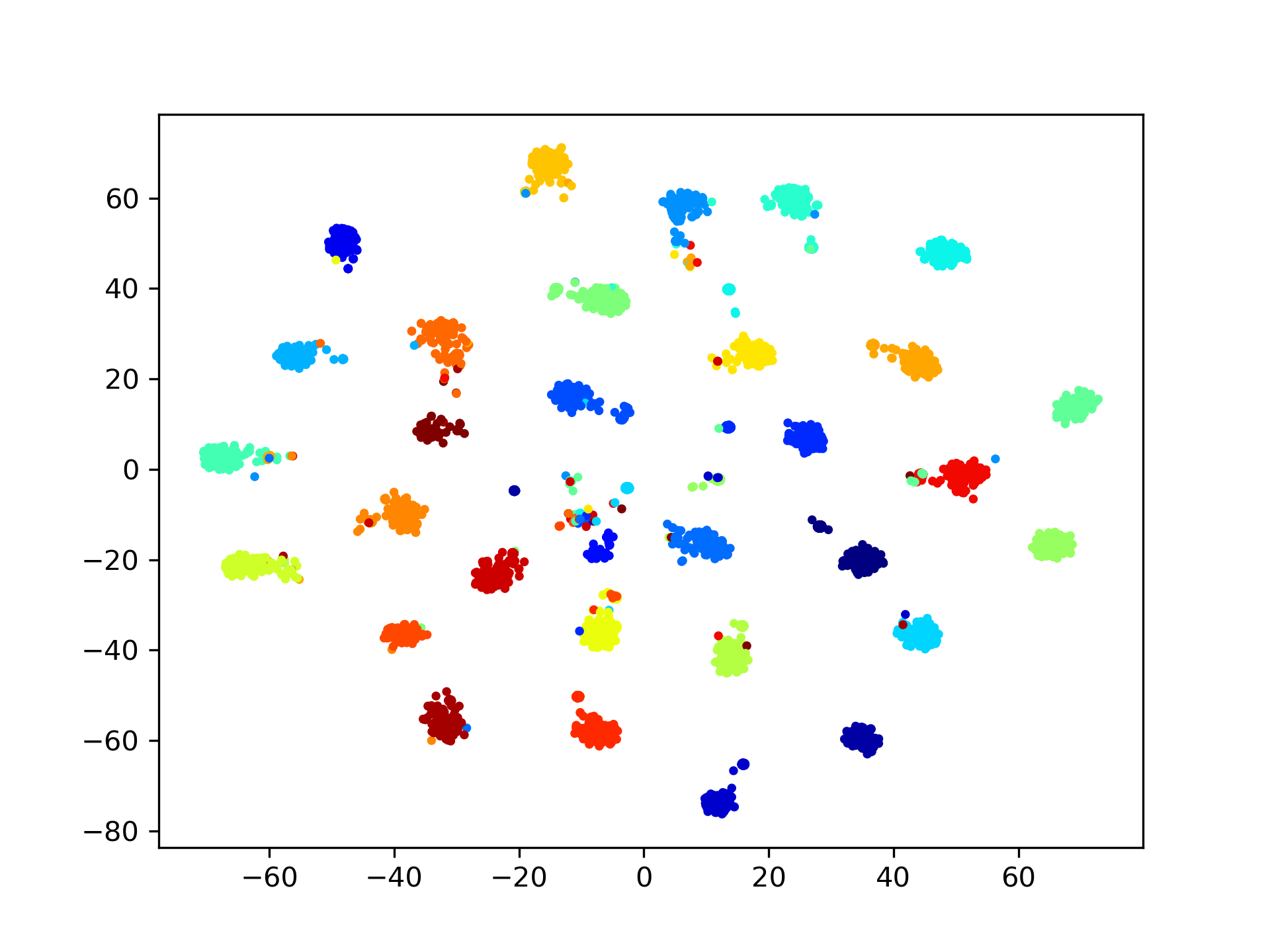}
	}
	\subfigure[Before training]{
		\centering
		\includegraphics[width=0.3\textwidth]{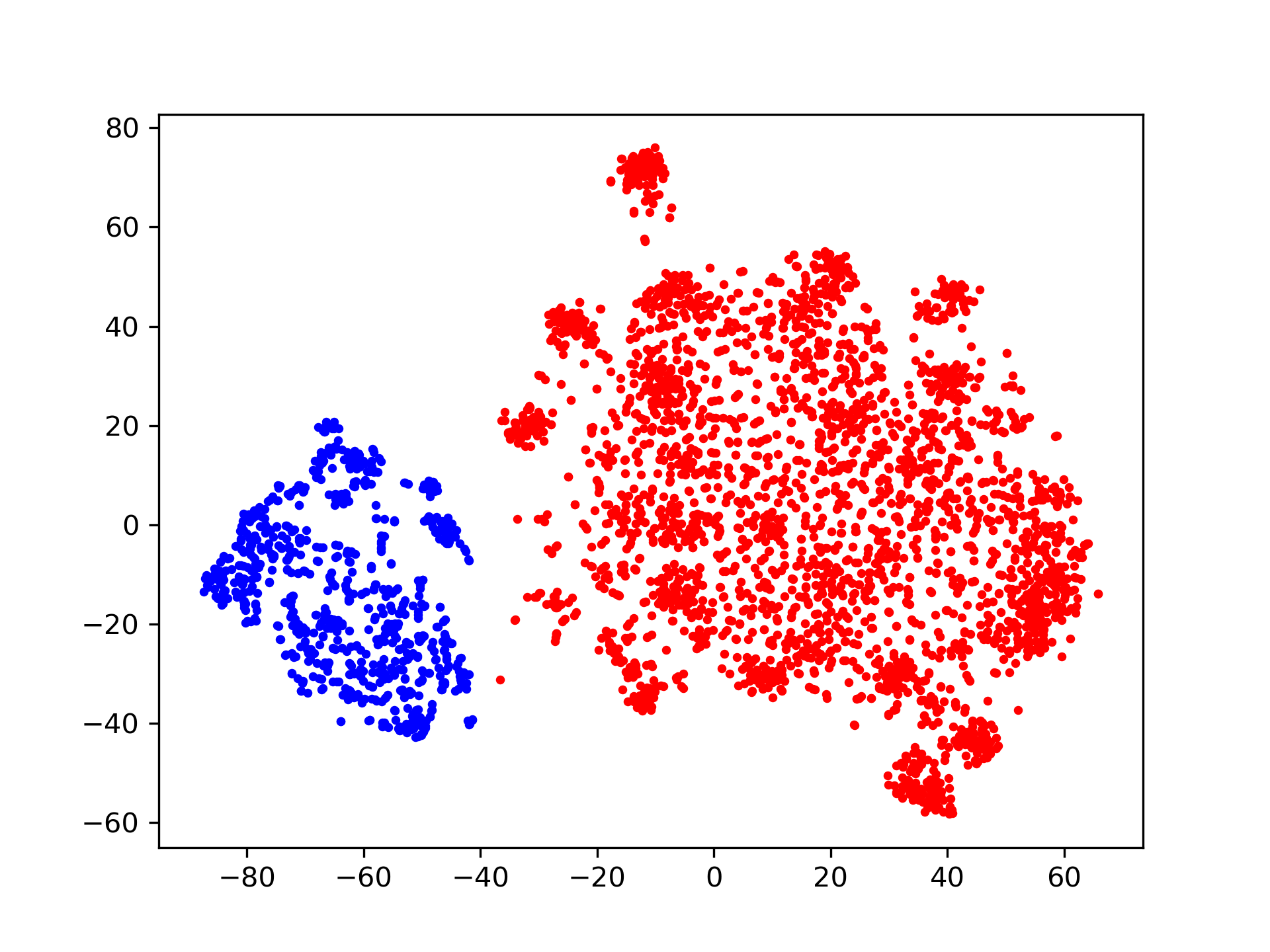}
	}
	%	\subfigure[ResNet-50]{
	%		\centering
	%		\includegraphics[width=0.22\textwidth]{./pic/a2wRD.png}
	%	}
	\subfigure[CDAN]{
		\centering
		\includegraphics[width=0.3\textwidth]{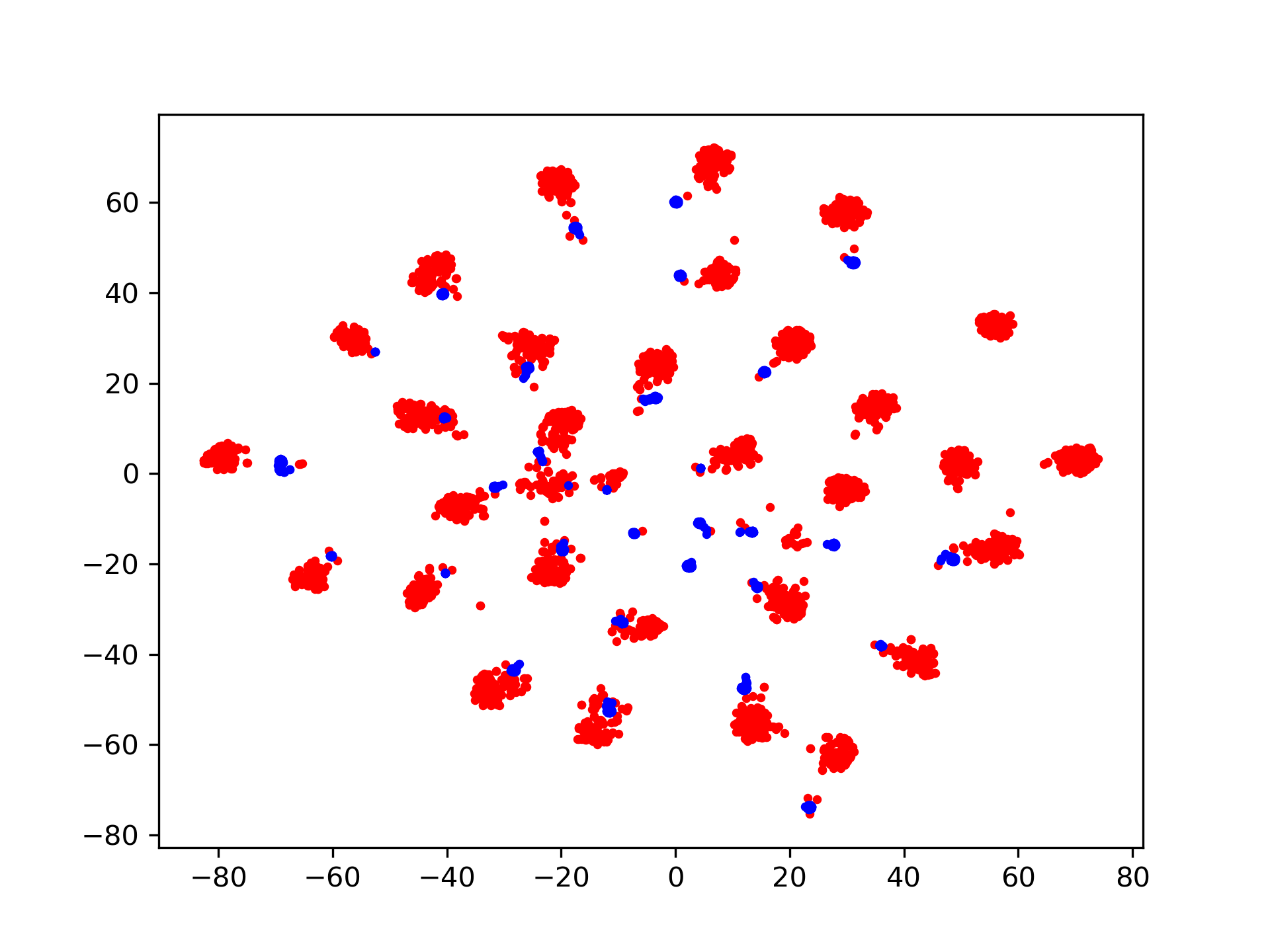}
	}
	\subfigure[REN]{
		\centering
		\includegraphics[width=0.3\textwidth]{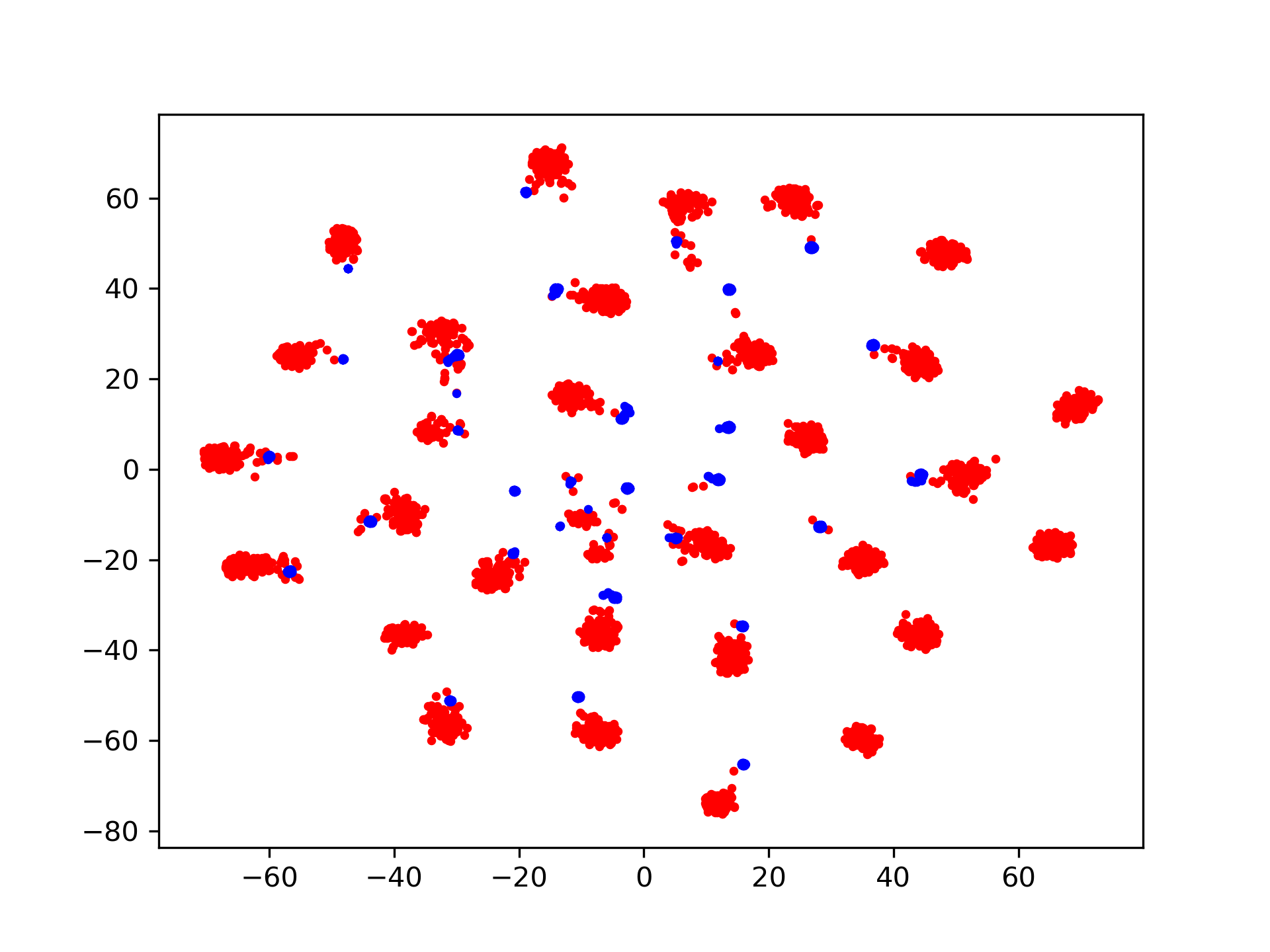}
	}
	\caption{The t-SNE visualization of A→W(Office-31). (a)(b)(c)represent category information and each color denotes a class.(d)(e)(f)Red circles are the source samples while blue circles are the target samples.}
	\label{fig4}
\end{figure}

To present the process of domain adaptation training more intuitively, we utilize the t-distribution Stochastic Neighbour Embedding (t-SNE) \cite{t-SNE} method to visualize the low-dimensional changes of features before and after adaptation in Fig. \ref{fig4}. We implement one task, namely A→W (Office-31) to perform these experiments. Obviously, before training, the spatial distribution of the source domain and target domain features is completely different. This indicates that the distribution contains no discernible intrinsic structure. Although after CDAN, most of data has been aggregated, there are still some classes indistinguishable. But after our method, the feature distribution shows a clear clustered structure. The cluster centers of the distribution are closer than before, and the degree of dispersion is more similar. This means that our method greatly improves the distribution in the feature space.

\section{Conclusion}

We propose a robust ensembling network for UDA based on model time ensembling and consistency constraint. It solves the negative transfer problem of target domain samples, which is close to the edge of the decision line due to adversarial learning. At the same time, the dual-network conditional adversarial loss proposed in this paper effectively decreases the instability in the adversarial learning process, and enables the network to learn more global transferable features. All-round experiments illustrate that our method is superior to the current mainstream methods on various domain adaptation datasets. %In future work, we are more interested in applying this method to other application scenarios of domain adaptation, such as object detection and image segmentation.

\bibliographystyle{splncs04}
\bibliography{References}
%
% \begin{thebibliography}{8}
% \bibitem{ref_article1}
% Author, F.: Article title. Journal \textbf{2}(5), 99--110 (2016)

% \bibitem{ref_lncs1}
% Author, F., Author, S.: Title of a proceedings paper. In: Editor,
% F., Editor, S. (eds.) CONFERENCE 2016, LNCS, vol. 9999, pp. 1--13.
% Springer, Heidelberg (2016). \doi{10.10007/1234567890}

% \bibitem{ref_book1}
% Author, F., Author, S., Author, T.: Book title. 2nd edn. Publisher,
% Location (1999)

% \bibitem{ref_proc1}
% Author, A.-B.: Contribution title. In: 9th International Proceedings
% on Proceedings, pp. 1--2. Publisher, Location (2010)

% \bibitem{ref_url1}
% LNCS Homepage, \url{http://www.springer.com/lncs}. Last accessed 4
% Oct 2017
% \end{thebibliography}
\end{document}